\documentclass[letterpaper, 10 pt, conference]{ieeeconf}  % Comment this line out if you need a4paper

\IEEEoverridecommandlockouts                              % This command is only needed if 
\usepackage{graphicx} % Graphics files. modern version of graphics
\usepackage{url} % Enable URLs

\usepackage{xcolor} % Text Color

\usepackage[symbol]{footmisc} % Custom Footnotes

\makeatletter
\newcommand\footnoteref[1]{\protected@xdef\@thefnmark{\ref{#1}}\@footnotemark}
\makeatother

\title{\LARGE \bf
Deploying the NASA Valkyrie Humanoid for IED Response: \\ An Initial Approach and Evaluation Summary 
}

\author{
Steven Jens Jorgensen$^{1,7,8}$, Michael W. Lanighan$^{2}$, Sylvain S. Bertrand$^{3}$,  Andrew Watson$^{1,4}$, \\ Joseph S. Altemus$^{1,4}$, R. Scott Askew$^{1}$, Lyndon Bridgwater$^{1}$, Beau Domingue$^{1,5}$, Charlie Kendrick$^{1,6}$, \\ Jason Lee$^{1}$, Mark Paterson$^{1,5}$, Jairo Sanchez$^{1}$,
Patrick Beeson$^{2}$, Seth Gee$^{2}$, Stephen Hart$^{2}$, \\ Ana Huaman Quispe$^{2}$, Robert Griffin$^3$, Inho Lee$^{3}$, Stephen McCrory$^{3}$, \\ Luis Sentis$^{7}$, Jerry Pratt$^{3}$, and Joshua S. Mehling$^{1}$
\thanks{The authors are with the $^{1}$NASA Johnson Space Center, $^{2}$TRACLabs, the $^{3}$Institute for Human Machine and Cognition (IHMC), $^{4}$Jacobs Technology, $^{5}$METECS, $^{6}$CACI, and $^{7}$The University of Texas at Austin.} \thanks{$^8$The author is partially supported by a NASA Space Technology Research Fellowship (NSTRF) Grant \# NNX15AQ42H}
\thanks{This work was supported by the Combating Terrorism Technical Support Office (CTTSO).}
}
\begin{document}

\maketitle
\thispagestyle{empty}
\pagestyle{empty}

%%%%%%%%%%%%%%%%%%%%%%%%%%%%%%%%%%%%%%%%%%%%%%%%%%%%%%%%%%%%%%%%%%%%%%%%%%%%%%%%
\begin{abstract}
As part of a feasibility study, this paper shows the NASA Valkyrie humanoid robot performing an end-to-end improvised explosive device (IED) response task. To demonstrate and evaluate robot capabilities, sub-tasks highlight different locomotion, manipulation, and perception requirements: traversing uneven terrain, passing through a narrow passageway, opening a car door, retrieving a suspected IED, and securing the IED in a total containment vessel (TCV). For each sub-task, a description of the technical approach and the hidden challenges that were overcome during development are presented. The discussion of results, which explicitly includes existing limitations, is aimed at motivating continued research and development to enable practical deployment of humanoid robots for IED response. For instance, the data shows that operator pauses contribute to 50\% of the total completion time, which implies that further work is needed on user interfaces for increasing task completion efficiency.\footnote{\label{fn:trademark_note}Disclaimer: Trade names and trademarks are used in this report for identification only. Their usage does not constitute an official endorsement, either expressed or implied, by the National Aeronautics and Space Administration.}
\end{abstract}

%%%%%%%%%%%%%%%%%%%%%%%%%%%%%%%%%%%%%%%%%%%%%%%%%%%%%%%%%%%%%%%%%%%%%%%%%%%%%%%%
\section{Introduction}
Humanoid robots are promising human avatars for dangerous tasks as they are kinematically able to traverse rough terrain, enter narrow passageways and tight spaces, manipulate objects, reach high places, and many other capabilities. Due to the potential of humanoid robots, the Combating Terrorism Technical Support Office (CTTSO) has an interest in exploring the use of humanoids for improvised explosive device (IED) response tasks. In other words, can a humanoid robot be used for bomb disposal?

\begin{figure}
\centerline{\includegraphics[width=1.0\columnwidth]{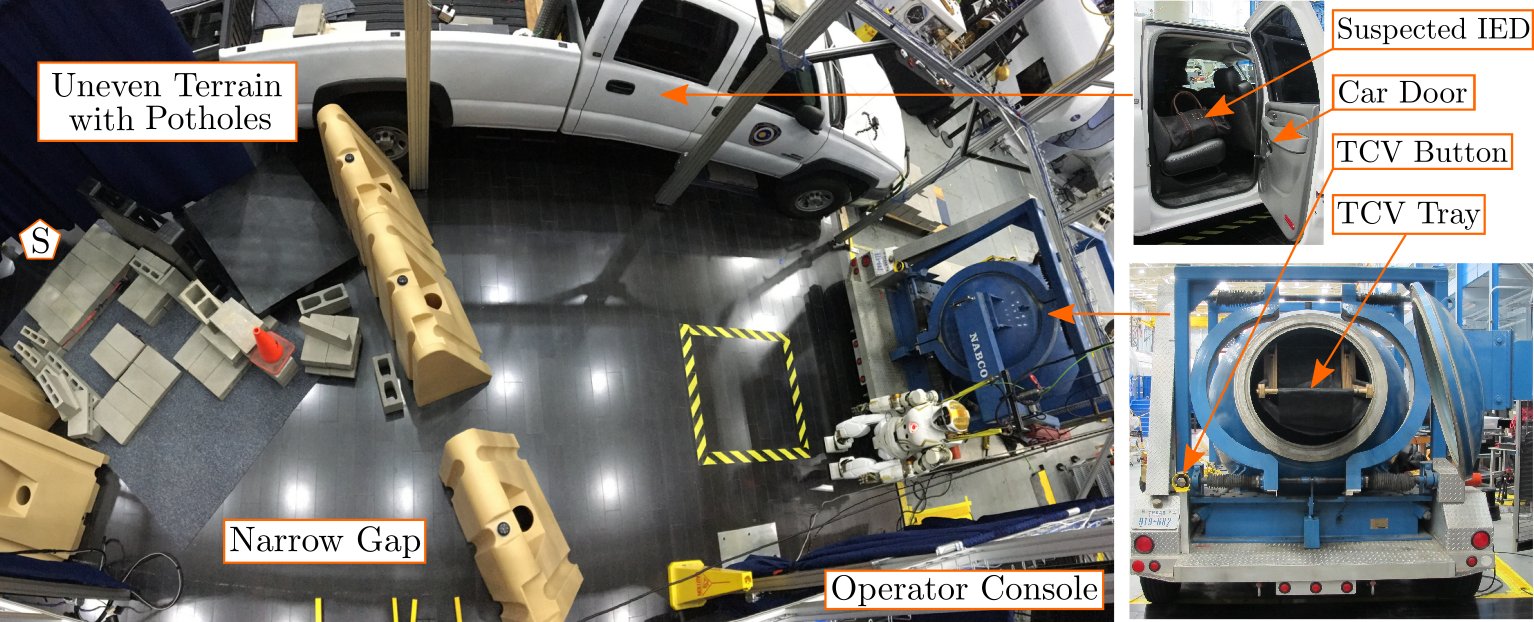}}
\caption{A panorama view of the IED response task field for feasibility testing. The humanoid begins at the area marked S. The task is to navigate the uneven terrain with potholes, cross a narrow passageway, open the car door, obtain the suspected IED, and secure it inside a TCV. The operator sits behind a curtain without direct visual access to the robot.}
\label{fig:eod_task_field}
\end{figure}

While a feasibility study could comprise of many different scenarios \cite{humphrey2009robotic}, specific bomb disposal tasks were chosen, in consultation with CTTSO and end users in the bomb technician community, that highlight the benefits of a humanoid form factor versus existing track-based robots \cite{yamauchi2004packbot, wells2005talon}. To this end, the NASA Valkyrie humanoid robot is tasked with traversing an uneven terrain with potholes, navigating through a narrow gap (representative of the space between parked cars), opening a car door, retrieving a suspected IED from within the car, and securing the IED inside a total containment vessel (TCV) (Fig.~\ref{fig:eod_task_field}). Note that a small tracked robot may be able to pass through a narrow gap, but it may have difficulty on uneven terrain, and such a platform would likely be too small to open a car door or operate a TCV. Similarly, a medium to large sized traditional bomb disposal robot may be able to traverse uneven terrain and accomplish all of the required manipulation tasks, but it would likely be too wide to pass through a narrow gap.

Multiple state-of-art technologies are leveraged to conduct this feasibility study and successfully perform a proof-of-concept demonstration. To have sufficient control but not overburden the operator during task execution, supervisory control over complete autonomy or teleoperation is preferred \cite{johnson2014coactive}. Concretely, the presented approach has similarities to \cite{johnson2015team, schwarz2017nimbro}, that are used in search, rescue, and disaster response scenarios \cite{negrello2018humanoids, krotkov2017darpa}. However, other novel tasks specific to IED response are considered, namely crossing a narrow gap, retrieving a suspected IED from a vehicle, depositing the IED in a TCV tray, and operating the TCV.  %While the DRC considered degraded network communications \cite{krotkov2017darpa}, this feasibility study did not. As the presented approach has similarities to those successfully deployed for the DRC, operating in degraded network communications is assumed to be viable. %\bt{Include comments about different search and rescue response tasks that uses human avatars.} 
%Emerging technologies for EOD robots also suggest that it's possible to maintain decent throughput for distances greater than 200m \cite{lowrance2017link}. 

Finally, certain restrictions are imposed for a successful feasibility demonstration. The robot must complete the end-to-end task in one hour. It must be fully operational until the IED is secured inside the TCV. It must not inadvertently engage the IED by, for instance, dropping it during transportation. The operator must also not have direct visual access to the robot.

%Additionally, while high-bandwidth radio frequency communication has line-of-sight limitations \cite{pezeshkian2007unmanned}, emerging technologies suggest that it's possible to maintain decent throughput for distances greater than 200m   \cite{lowrance2017link}. 

%As this work have similar goals of the DARPA Robotics Challenge (DRC) \cite{krotkov2017darpa} to deploy capable robots in hazard or disaster-stricken human environments

%Existing EOD robots \cite{yamauchi2004packbot, wells2005talon, kilitci2011analysis} have been successfully deployed for thousands of hours, but these robots exclusively have track bases that have a number limitations. To name a few, its mobility is restricted to the surface area of the base. While small bases can fit in tight spaces, it cannot have large arms or carry heavy payloads. 

% \begin{figure}
% \centerline{\includegraphics[width=\columnwidth]{media/front_page.png}}
% \caption{A snapshot of the the Valkyrie robot preparing to drop the bag containing an improvised explosive device (IED) in the total containment vehicle (TCV) tray. As the tray is 154.3cm off the ground, Valkyrie's high reaching capability enables successful task completion}
% \label{fig:frontpage_bagdelivery}
% \end{figure}

%While the DARPA Robotics Challenge (DRC) Trials and Finals highlighted some of the tasks that robots need to perform in place of humans, there are additional tasks that an IED robot traversing a narrow passage-way, opening a car door, and operating a total containment vehicle.

\begin{figure*}
\centerline{\includegraphics[width=1.75\columnwidth]{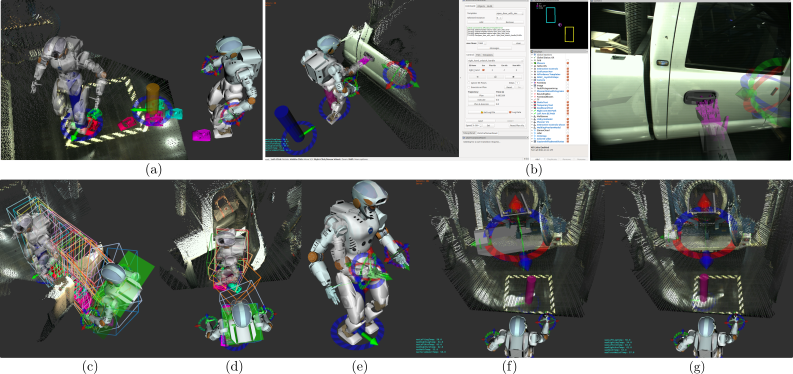}}
\caption{A visualization of some of the enabling technologies for IED response. (a) shows the planned footsteps for multiple waypoints. (b) shows the operator console view during the door opening task. A 6 DoF interactive marker is used to control the hinge location of the door and a transparent arm shows a preview of the manipulation trajectory. On the right, the operator can toggle different visualization modes, observe a top-view of the center-of-mass and support polygon, and observe a camera view with affordance template overlays. (c) and (d) show the different bounding box approximations used for crossing the narrow gap and entering the car door. (e) shows the online IK solver previewing the arm configuration for a user-specified end-effector pose. Finally, (f) and (g) show the TCV bumper and tray affordance templates. The purple marker indicates where the robot should stand to safely raise an object. In (f), the user roughly aligns the template to the point cloud, while in (g) automatic object registration was used to snap the template in place. }
\label{fig:enabling_tech_images}
\end{figure*}

\section{Enabling Technologies for Humanoid IED Response}

\textbf{The NASA Valkyrie Humanoid Robot} \cite{radford2015valkyrie} has 44 degrees of freedom: seven degrees in each arm, six degrees in each leg, three in the torso, three in the neck, and six in each of the four-fingered hands. The arm, torso, and leg have series elastic actuators (SEA) that enable torque control with a bandwidth of up to 70Hz and a resolution of $2\cdot 10^{-3}$N$\cdot$m)\cite{paine2015actuator}, while the neck is position controlled, and tendon-driven fingers are current controlled. In addition to absolute position encoders, incremental encoders, and spring deflection sensors in each SEA, the robot uses a Microstrain inertial measurement unit (IMU) sensor on both the pelvis and torso, an ATI force-torque (F/T) sensor in the sole of each foot for center-of-pressure (CoP) estimation, and a Multisense sensor for LIDAR and stereo data.

\textbf{Whole-body Controller for Walking and Manipulation.}
The robot uses a momentum-based whole-body controller that is framed as a quadratic program (QP) \cite{koolen2016design}. A set of task space accelerations and external forces is fed as objectives to the QP while attempting to track a desired rate of change of centroidal momentum and minimize joint accelerations and contact forces. This QP outputs a joint acceleration vector and contact wrenches which are used to calculate the desired actuator torques using inverse dynamics. The walking behavior is based on the ideas of Capture Point (CP)\cite{koolen2012capturability}. Concretely, the instantaneous capture point (ICP) is planned using  \cite{seyde2018inclusion, englsberger2015three}, which is then used to compute the desired momentum rate \cite{griffin2017walking} for the QP. This CP-based walking has useful stability properties that are used to monitor the safe execution of trajectories and pause potentially unsafe walking trajectories. Additionally, a variety of high-level interfaces\footnote{https://github.com/ihmcrobotics/ihmc-open-robotics-software}, such as task-space and joint-space spline trajectory generation, planning toolboxes, and control mode changes are exposed to aid with commanding the robot.

\begin{figure*}
\centerline{\includegraphics[width=1.75\columnwidth]{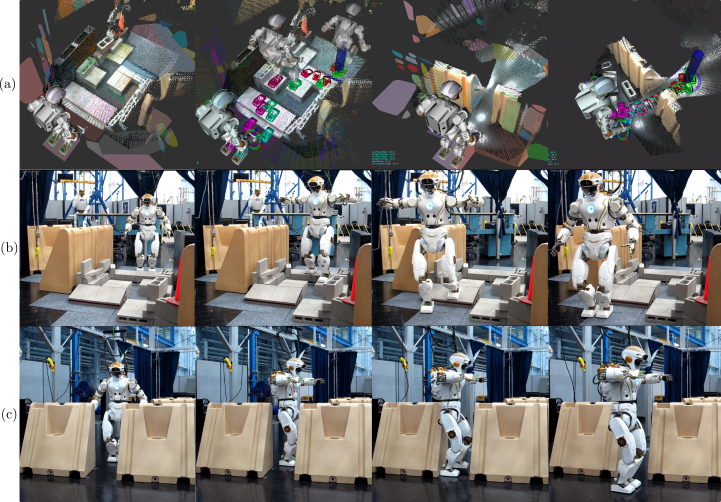}}
\caption{A collage of the main locomotion tasks. The top row of subfigures (a) show the operator's view with the superimposed planar regions and the planned footstep paths for both the uneven terrain task and the narrow gap task. The footstep planner is able to snap desired footsteps onto the detected planar regions, avoid unsafe footholds such as potholes, and plan paths through a narrow passageway. The middle and bottom rows (b) and (c) subfigures show a snapshot sequence of uneven terrain traversal and narrow gap crossing respectively.}
\label{fig:locomotion_collage}
\end{figure*}

\textbf{Waypoint Navigation with Environment Awareness.}
An operator sends locomotion commands primarily by placing a waypoint to indicate a desired goal destination for the robot. Additional waypoints can also be added to aid the planner with breaking down complex footstep plans (Fig.~\ref{fig:enabling_tech_images}a). Terrain data is obtained using the aggregate of the LIDAR point cloud data over a specified time window. Then, a planar detection algorithm based on \cite{marion2016perception} is used to segment traversable planar regions in the environment. Given the current stance of the robot and the waypoints, an A* based planner inspired from \cite{garimort2011humanoid, hornung2013search}, is used to find the sequence of footsteps to reach the desired end goal (Fig.~\ref{fig:locomotion_collage}a). To address collisions when exploring the A* graph, a scalable 3D bounding box is used to approximate the robot and collision checks are queried between the bounding box and the segmented planar regions  (Figs.~\ref{fig:enabling_tech_images}c,~\ref{fig:enabling_tech_images}d). The output of the planner is a sequence of footsteps that are collision free assuming an appropriate bounding box approximation. 

% \textbf{Operator Situational Awareness and Planner Feedback Visualization}
% One of the key complaints EOD technicians from CTTSO voiced was the lack of situational awareness due to the limited camera view of existing EOD robots. The combination of using automatic planar segmentation described above and the use of a colored Octomap \cite{hornung13auro} helped alleviated this problem during the feasibility tests. 
% The operator is also able to visualize the robot's intent by being able to preview the walking path of the robot as well as the proposed trajectory plans of the robot.

\textbf{Affordance Templates (AT) and State Machines for High-Level Execution of Goals.}
%Certain tasks such as opening car doors, manipulating bags, and operating a TCV will be repeatedly required from an EOD robot. To this end, a
An affordance template (AT) \cite{ hart2015affordance} is used to encode strategies to manipulate objects via a series of end-effector waypoints and stance locations defined in the frames of those objects. Once an AT has been constructed, it can be reused for similar manipulation tasks. To use an AT, the operator registers/aligns a parameterizable representation of the object to the point cloud data, enabling the robot to execute manipulation trajectories on the object. Additionally, state machines are used to encode common sequences of operations and decision making processes. Combining ATs and State Machines enables an operator to achieve desired manipulation goals by only sending high-level commands to the robot \cite{beeson2016cartesian}.

%is a graphical construct that enables an operator to adjust robot task goals. It contains manipulation trajectory information by way of storing a series of end-effector waypoints defined in the object-centered frame.  are human-adjustable graphical constructs 

\textbf{Automatic Object Registration}
As aligning affordance templates can be time-consuming for the operator, an object registration scheme was created to automatically fit known object mesh models to the LIDAR point cloud data. The registration scheme is similar to the standard iterative closest point algorithm \cite{rusinkiewicz2001efficient}. However, the minimization error is formulated as a non-least squares optimization problem that is fed into the Ceres solver\footnote{Ceres:\url{http://ceres-solver.org/}}. This has the advantage of using quasi-Newton gradient descent approaches that are better at avoiding local minima. %Additionally, gradient descent is performed with $N$ error signals, where $N$ is the number of 3D points in the source cloud, which allows it to better find the global minima. 

\textbf{Task-Space Control and Stored Poses.}
Existing planner limitations still necessitate some level of task-space control. This functionality is particularly necessary when the locomotion or manipulation planners fail. For Valkyrie's hands, a fast online IK solver \cite{beeson2015trac} is used to preview potential arm configurations given user-specified end-effector poses \cite{johnson2014coactive}. For Valkyrie's feet, a sequence of user-specified footstep landing configurations can be previewed and executed. Sliders for the pelvis location and torso orientation are also available to further exploit whole-body configurations. Finally, the user can also load stored poses for common desired configurations.

%\textbf{ROS-centered middleware communication}
%The Robot-Operating System ROS \cite{quigley2009ros} is primarily used to facilitate communication between processes and establish input and output protocols. The IHMC controller interface extensively used ROS  

\section{IED Response Task Scenario Description}
\textbf{Traversing Uneven Terrain and a Narrow Gap.} The robot's starting location (marked S in Fig.~\ref{fig:eod_task_field}) is in front of a series of stacked and angled cinder blocks used to simulate uneven sidewalks, inclines, and potholes commonly encountered by bomb technicians in urban environments. To simulate passing between parked cars, a narrow gap is constructed using a pair of jersey barriers that can be placed between 44 and 51 cm apart.

\textbf{Retrieving a Suspected IED from a Parked Vehicle.} After traversing the uneven terrain and narrow gap, the robot must retrieve a bag containing a suspected IED from inside a parked vehicle (Fig.~\ref{fig:eod_task_field}).

\textbf{Securing the IED inside the TCV.} Once the IED has been retrieved, the robot must safely transport it to the TCV and secure it inside. Securing the IED involves depositing it inside the TCV tray, pushing the TCV tray inside the vessel, and pushing the TCV close button.

\begin{figure*}
\centerline{\includegraphics[width=1.7\columnwidth]{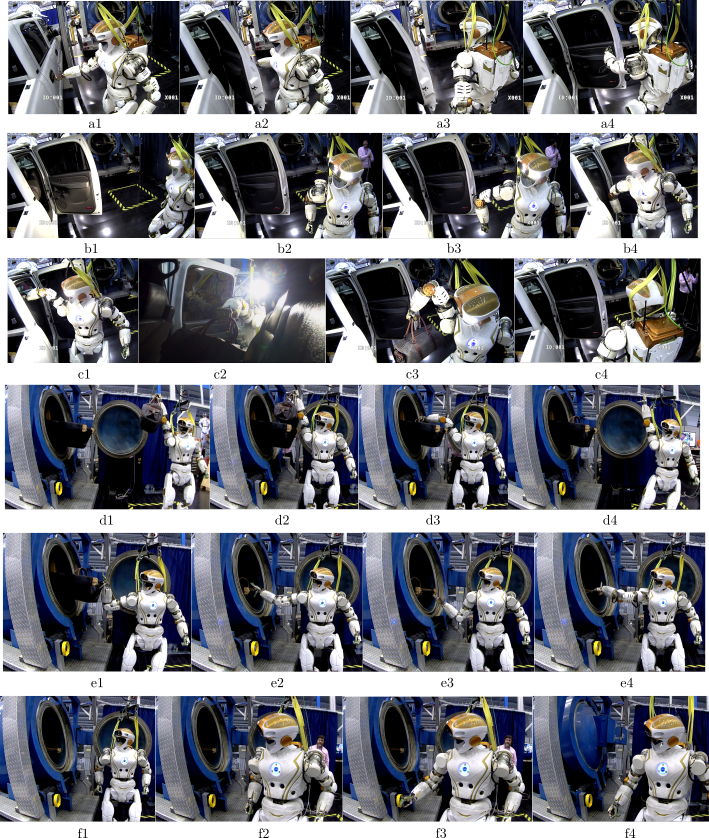}}
\caption{A collage of all the IED response manipulation tasks showing (a) the door opening, (b) the door entry, (c) the bag retrieval, (d) the TCV bag placement, (e) the TCV tray push, and (f) the TCV button push sequences. Please see Sec.~\ref{sec:execution_approach_and_discussion} for a detailed description of the manipulation sequences.}
\label{fig:manipulation_collage}
\end{figure*}

\section{Execution Approach and Discussion}
\label{sec:execution_approach_and_discussion}
\textbf{Traversing Uneven Terrain.}
\label{sec:approach_uneven_terrain}
After planar regions have been segmented, the operator loads a configuration file to update the footstep planner with parameters suitable for uneven terrain planning. Then, a 2D waypoint is placed at the end of the uneven terrain field. The planner can take between 10 to 40 seconds to return a footstep plan. The operator previews the walking result and executes the walking trajectory if the preview looks appropriate (see Fig. \ref{fig:locomotion_collage}a). Fig. \ref{fig:locomotion_collage}b shows a sequence of snapshots of Valkyrie traversing the uneven terrain.

\begin{table*}[t]
\caption{Successful End-to-end Task Completion Times for each IED Response Sub-task. }
\label{table:task_completion_times}
\begin{center}
\begin{tabular}{|c|c|c|c|c|c|}
\hline
IED Response        & Best run    & Best run time w/out    & Operator paused time (s)    & Min. time from       & Time average of \\
Sub-tasks  &  time (s)    & operator pauses (s)    & during the best run         & successful runs (s)   & successful runs (s) \\
\hline
Uneven Terrain Traversal & 99  & 57  & 42 (42.42\%) & 79 & 132  \\
\hline
Narrow Gap Crossing      & 180 & 94  & 86 (47.78\%) & 152 & 277 \\
\hline
Car Door Opening         & 210 & 120 & 90 (42.86\%) & 210 & 288 \\
\hline
Car Door Entry           & 149 & 83  & 66 (44.30\%) & 141 & 206 \\
\hline
Bag Retrieval            & 180 & 94  & 86 (47.78\%) & 124 & 295 \\
\hline
Walk to TCV Area         & 151 & 67  & 84 (55.63\%) & 125 & 168 \\
\hline
TCV Bag Drop-Off         & 214 & 122 & 92 (42.99\%) & 214 & 298  \\
\hline
TCV Tray Push            & 261 & 119 & 142 (54.41\%) & 152 & 384 \\
\hline
TCV Button Push          & 165 & 59  & 106 (64.24\%) & 139 & 213 \\
\hline
\hline
Total Time (s)               & 1609 & 815 & 794 (49.35\%) & 1336 & 2262 \\
\hline
Total Time (mins + s)        & 26 mins + 49s & 13 mins + 35s & 13 mins + 14s & 22 mins + 16s & 37 mins + 42s \\
\hline
\end{tabular}
\end{center}
\end{table*}

In testing different types of cinder block configurations, the following problems arose. Certain environment configurations can create perceptual occlusions which prevent a traversable plane or foothold from being detected. Next, predefined footstep swing trajectories can cause the robot's toe to stub on the cinder blocks during step-up trajectories or its heel to hit cinder blocks during step-down trajectories. Fast swing trajectories can saturate the leg joint velocities resulting in tracking errors which in turn cause inaccurate foot landing locations and rough touchdowns that degrade state estimation.  For difficult terrains, state estimation drift on the scale of 10cm can be observed, which reduces the safety of footstep executions. Finally, due to actuator torque saturation and controller limitations, there is a maximum z-height footstep that the robot can take unless multi-contact locomotion is exploited.

\textbf{Crossing a Narrow Gap.}
Similar to the previous navigation task, a 2D waypoint is dropped on the far side of the narrow gap to obtain a footstep plan. However, to safely execute the task, the operator has to also perform the following operations. First, the planar region buffer is reset to obtain new planar region estimates and alleviate the accumulated state estimation drift from the previous task. Then, to address footstep planner limitations, the operator raises the pelvis (thereby straightening the legs) to decrease the depth occupancy of the robot, raises the arms to further minimize the occupancy volume below the waist level, and loads an appropriate-sized bounding box for the footstep planner. Fig.~\ref{fig:enabling_tech_images}c shows how the bounding box approximation crosses the narrow gap. Without these additional operator actions, arm and knee collisions with the jersey barrier can occur, and operator intervention and replanning will be needed. Note also that an overly restrictive bounding box will prevent the planner from finding a footstep plan. Otherwise, Fig.~\ref{fig:locomotion_collage}a and \ref{fig:locomotion_collage}c show a successful footstep plan and a snapshot sequence of a narrow gap crossing. 

\textbf{Opening the Car Door.}
After crossing the narrow gap, the pelvis height is reset to default. The operator aligns an AT for the door (Fig.~\ref{fig:enabling_tech_images}b). After alignment, a stored 2D waypoint w.r.t. the door handle indicates the ideal stance location for opening the car door. After navigating to the stance location, a state machine is executed to open the car door in two stages. First, the door handle is unlatched and pulled to the first detent (Fig.~\ref{fig:manipulation_collage}a1, \ref{fig:manipulation_collage}a2). Next, the robot navigates to a new stance location and uses its left arm to fully open the door (Fig.~\ref{fig:manipulation_collage}a3, \ref{fig:manipulation_collage}a4). During the state machine execution, it prompts the operator whether it is safe to continue the door opening trajectory or not. It is possible for the task to fail when grasping the handle, removing the fingers from the handle, or if the door is accidentally closed by a knee collision during the second stage of door opening. If such failures occur, the operator needs to intervene and repeat the AT alignment and execution process.

The door opening task revealed that the stance location of the robot is critical for reachability and collision considerations during manipulation. Currently these stance locations are identified a priori. Notice also that the robot is either manipulating or locomoting but never doing both at the same time. This limitation can be apparent for spring loaded doors with no detents, which will require coordinated whole-body movements for task completion. 
%\bt{It remains future work to determine how much of the AT and state machine is reusable for different car doors. Reconfiguring and/or augmenting the AT to include other relevant information such as forces or impedances is also currently being investigated}. An online procedure for updating template parameters during task execution \bt{(such as in \cite{arduengo2019versatile})}, \bt{could potentially aid in creating} robust and generic ATs.

\textbf{Entering the Car Door.}
Due to state estimation drift during walking, the operator first clears the planar region buffer and retreats from the door to segment new planar regions (Fig.~\ref{fig:manipulation_collage}b1). The robot is then repositioned, and a stored pose for the right arm is loaded to enter the car with a leading right arm (Fig.~\ref{fig:manipulation_collage}b2 - \ref{fig:manipulation_collage}b4). A different bounding box approximation (Fig.~\ref{fig:enabling_tech_images}d) is set in the footstep planner and a 2D waypoint navigation encoded in the AT is used to safely enter the car door.

\textbf{Retrieving the IED Bag.}
Once the robot enters the car, an AT with approach-and-grasp waypoints is aligned to grab the bag handle (Fig.~\ref{fig:manipulation_collage}c1,~\ref{fig:manipulation_collage}c2). Visual inspection and regrasp procedures are used to ensure a solid grasp. Notice that a strong assumption is made that the bag is in a location that is within the robot's reach. If the bag was located on the car's floorboard or deeper into the car, multi-contact approaches will be needed to enable the robot to lean in or mount the vehicle to obtain the bag.

\textbf{Walking to the TCV Area to Deposit the IED.}
After retrieving the bag, the robot needs to exit the car door and walk to the TCV area (Fig.~\ref{fig:manipulation_collage}c3,~\ref{fig:manipulation_collage}c4). Similar to the planner limitations exposed by the narrow gap crossing task, it is still possible to hit the car door upon exiting. If the car door is hit, this is an unmodeled disturbance from the controller's perspective. Upon detecting large tracking errors, the robot is able to halt its walking trajectory autonomously, adding operational safety, and enabling the operator to intervene and plot manual footsteps for recovery and egress. Once a successful egress is performed, the operator navigates by setting a series of waypoints to find a visual of the TCV and aligns a TCV AT to the point cloud data. A mesh model is used to automatically snap the AT in place (Fig.~\ref{fig:enabling_tech_images}f,~\ref{fig:enabling_tech_images}g).

\textbf{TCV Bag Drop-Off.}
Next, the operator navigates to a stance location defined w.r.t the TCV bumper AT to safely raise the bag (Fig.~\ref{fig:manipulation_collage}d1). To clear the tray's height, the pelvis is raised. The operator then navigates to a stored stance location w.r.t. the TCV tray (Fig.~\ref{fig:manipulation_collage}d2). The bag drop-off is then done with a sequence of stored poses (Fig.~\ref{fig:manipulation_collage}d3). Finally, the arm is raised, and the operator commands a retreat from the TCV tray to create enough space to safely bring the arm back down (Fig.~\ref{fig:manipulation_collage}d4). The combination of lateral upper body movements when walking and a raised arm configuration can cause the robot to hit the tray when retreating, which is sometimes enough disturbance for the robot to halt its walking trajectory. Thus, there are instances that the robot cannot retreat without significant operator intervention.
%\bt{Of note, slow swing foot trajectory times result to walking trajectories that approach quasi-static behaviors which can cause large lateral upper body movements. Because the preferred swing time parameter is slow to improve trajectory tracking, the combination of large lateral upper body movements and a raised arm configuration can cause the robot to hit the tray when retreating, which is sometimes enough disturbance for the robot to halt the walking trajectory. Thus, there are instances that the robot cannot retreat without significant operator intervention.}

% Large upper body swinging motions are due to having a preferred slow swing foot time
% Since the robot swings its upper body during locomotion, the raised arm configuration can hit the tray when retreating causing enough disturbance for the robot to halt the walking trajectory. \bt{There are instances that the robot cannot retreat without significant operator intervention. O

%. During the bag drop off sequence, it is possible for the robot to fail to retreat from the TCV tray \bt{without significant operator intervention}

\textbf{TCV Tray and Button Push.}
To secure the IED bag, the TCV tray needs to be pushed inside the vessel, and the TCV needs to be closed with a button push. First, the operator navigates to a stored stance location. Then the pelvis is shifted away from the TCV to allow for an overhand approach trajectory with the TCV tray AT (Fig.~\ref{fig:manipulation_collage}e1). Stored poses are used to move the pelvis and right arm to push the tray inside the vessel (Fig.~\ref{fig:manipulation_collage}e2). The TCV tray AT is realigned by the operator and a final underhand push trajectory using the AT is deployed which completes the tray push task (Fig.~\ref{fig:manipulation_collage}e3,~\ref{fig:manipulation_collage}e4). Finally, the arms are tucked and the operator uses the TCV button AT to navigate to a stored stance location and deploy the push button trajectory (Fig.~\ref{fig:manipulation_collage}f). The entire trajectory can be performed using a state machine with user prompts that allow the operator to intervene if necessary. While the entire maneuver is complex, ATs improve task performance and reduce operator burden, in large part because the TCV is a known object that can be modeled a priori.

\begin{table}[tp]
\caption{ Feasibility Run Classifications}
\label{table:run_classifications}
\begin{tabular}{|c||c|c|}
\hline
\textbf{Classification of 33 runs}  & \textbf{Count} & \textbf{Count/Total Runs}  \\
\hline
Successful Runs & 19 & 57.57\% \\
\hline
Failed Runs due to falls & 10 & 30.30 \% \\
\hline
Failed Runs due to an unrecoverable  & 4 & 12.12\% \\
 walking configuration               & & \\
\hline
\hline
\textbf{Fall Types } & \textbf{Count} & \textbf{Count/Total Runs} \\
\hline
Cinder Block Toe Stubs & 4 & 12.12\%   \\
\hline
State Estimation Drift & 2 & 6.06\%   \\
\hline
Balance Controller Failure & 2 & 6.06\%  \\  
\hline
Footstep Planner Implementation Error & 1 & 3.03\%   \\
\hline
Aborted Walking Trajectory & 1 & 3.03\%  \\
\hline
\end{tabular}
\end{table}

\section{Feasibility Test Runs and Results}
After developing minimum viable approaches for each task, a number of timed runs were conducted to assess performance in simulated operational conditions. These runs were helpful in identifying missing capabilities and in directing further development efforts. While the robot hardware and software configuration varied from run to run, developers used a best-effort configuration for each timed run. This loosely models how the system would be used in an actual real-world deployment. Once task completion time fell below one hour, the physical environment was set constant so that consistent metrics could be used to assess performance and identify limitations of the described approach. Tables~\ref{table:task_completion_times} and ~\ref{table:run_classifications} show quantitative data for all timed runs once end-to-end completion times of less than one hour were achieved.

\begin{figure*}
\centerline{\includegraphics[width=1.35\columnwidth]{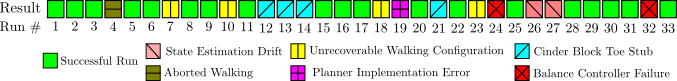}}
\caption{A visualization of the feasibility runs in the order it was performed and their result classification (best viewed in color).}
\label{fig:run_classification_visualization}
\end{figure*}

%While different types of cinder block configurations and varying the narrow gap distances were attempted, the bag, car, and TCV locations were kept constant. The first time an end-to-end task demonstration was accomplished took 3 hours with the largest slowdowns occurring from all the manipulation tasks. Only after identifying the subtleties of each task and making improvements on the AT's manipulation waypoints and ideal stance locations did the overall task completion time come down to an average of \rt{35 mins}. It was at this point that useful metrics can be obtained regarding robot reliability, operator throughput, as well as the true limitations of the approach under best case scenarios.

Table~\ref{table:task_completion_times} catalogs successful task completion times. The first column breaks down the entire scenario by sub-task. The second column lists sub-task completion times for the best end-to-end run. The third column lists the total time that the robot was in motion (no operator pauses) during the best run. The fourth column is the total operator paused time during the best run, which represents the total time that the robot was stationary due to operator initiated pauses or operational delays (e.g. waiting for plans, confirming plans, changing control modes, rotating camera views, or manipulating interactive markers to prepare the robot for motion). The fourth column also illustrates the percentage of task time resulting from operator-related delays. Among all successful runs, the fifth column lists the minimum completion time for each individual sub-task. The final column catalogs the time average across all successful runs. For an end-to-end run time of 26 mins and 49s, the robot moved continuously for 13 mins and 35s, providing a lower bound time for task completion. However for the remaining 13 mins and 14s, the robot waited for an operator command. Even under best case scenarios, 50\% of the task completion time is spent on operator pauses with the current approach. It is noteworthy that the manipulation tasks were the most time consuming portion of the scenario. Only after identifying the subtleties of each task and making improvements to the AT's manipulation waypoints and ideal stance locations did the overall task completion time drop to an average of 37 minutes.

Table~\ref{table:run_classifications} classifies the results of the feasibility runs, and Fig.~\ref{fig:run_classification_visualization} provides a visualization of the same data displayed in the order that each run was performed. Out of 33 timed runs, 19 were successful, 10 failed due to robot falls, and four failed due to unrecoverable walking configurations. It is noteworthy that successes and failures of similar types typically occur in close proximity to one another. Of 14 failures, 6 can be attributed to incorrect F/T readings leading to falls (See Sec.~\ref{sec:limitations_discussions}, \textbf{Addressing Hardware Issues}). Four toe-stub related falls on the cinder blocks were due to predefined swing trajectories and state estimation drift. Two falls were directly due to state estimation drifts. In one of these, the robot attempted a foothold that was no longer safe. In the other, the stored planar regions had drifted after a few footsteps causing future plans using old planar regions to include footstep landing locations above the ground plane. 

One robot fall was caused by a balance controller failure when computed NaN values were propagated to the controller by external trajectories with very small time intervals. Another balance controller related fall was due to accumulated debris on the foot causing the robot to lose sufficient contact friction with the ground. There was also a planner implementation failure in which the footstep plan contained previously traversed footsteps. Finally, one fall was caused by aborting the walking trajectory after the robot paused from hitting the TCV bumper. This exposed a software bug that did not update the number of contact points, which lead to an inconsistency between center of mass trajectory planning and the active foot support region.

Failures with an unrecoverable walking configuration occurred when there was either no sequence of footsteps to safely egress out of tight spaces (due to unavoidable body collisions with the environment), or when the robot's walking controller indefinitely paused the walking trajectory by refusing to perform a contact transition due to a combination of large tracking errors, the dragging of the safety hoist, inaccuracies in the robot model, and the leg approaching singularity. While the former only occurred during development, the latter occurred four times during timed-runs. Pausing the walking trajectory is primarily a safety feature to prevent falls when tolerance values for safe contact transitions are not met.  While this is indeed a useful safety feature during impacts with the environment, for example, it can be a hindrance when the robot pauses its walking trajectory in free space. If this occurs, overwriting the pause command is usually safe, but the runs were aborted to debug the issue.  

While the causes of failures have been addressed since the completion of the feasibility study, the quantitative data illustrates the difficulties of deploying robust bipedal locomotion. However, it is also important to note that robustness to these failure types are possible in the future. 

%Most of the run failures are primarily due to robot falls that occurred on the uneven terrain traversal which were discussed in Sec.~\ref{sec:approach_uneven_terrain}. As the robot cannot recover from a fall, the EOD task cannot be completed. It is necessary that the robot be both robust to and able to recover from falls for any EOD missions.

%A few robot falls did occur when egressing from the car door and approaching the TCV with a bag. These falls were primarily due center-of-pressure estimation drift from the thermal expansion of the modified ATI sensor which deteriorates the walking behavior and further causes state estimation drifts. This also lead to the drifting of segmented planar regions causing the robot to land footsteps in the air.

\section{Additional Limitations, Discussions and Future Directions}
\label{sec:limitations_discussions}
%In addition to the discussions in Sec.~\ref{sec:execution_approach_and_discussion}, this section expands on other identified limitations of the current approach as well as provide directions on future research and development.

%\textbf{Improving user interface.}
%A constant problem during robot operation is the high number of operator mouse clicks needed to setup the robot for task execution. For instance, the operator can only select one part of the gimbal at a time to manipulate any interactive object in the console such as an AT, the robot's end-effectors, or the waypoint markers. While utilizing a mouse enables flexibility due to its ability to open context menus, switch panels, rotate views, and drag markers, its serialized interface limits the throughput of operator commands. Utilizing VR-related technology, \bt{e.g.} \cite{allspaw2018remotely}, is actively being pursued to increase throughput and enable dexterous manipulation.

\textbf{Improved User Interfaces.} The user interfaces to operate Valkyrie are currently based on the ROS RViz environment with adjustable markers \cite{gossow2011interactive} and custom panels. While these tools provide a high-level of flexibility, they are inevitably limited when used to control a humanoid operating in a 3D environment as they rely on 2D computer screens and a mouse and keyboard. Moreover, such interfaces are not currently optimized for use in the field as it can have a serialized interface implementation which limits the throughput of operator commands. The use of VR-related technology \cite{allspaw2018remotely} is actively being pursued to increase throughput and improve dexterous manipulation performance.   

\textbf{Further exploiting Whole body trajectories.}
Despite having a humanoid robot, the approach presented here largely separates locomotion from manipulation, which limits the overall capabilities of the robot. For example, performance during the IED retrieval and tray push tasks would likely benefit if additional contacts could be used to lean against the environment. This ability would also provide greater adaptability to other retrieval scenarios. Therefore, exploiting multi-contact motions \cite{chung2015contact} in the future is highly desirable as it would enable ladder climbing, crawling, reaching further into confined places, etc.

\textbf{Incorporating Localization.}
Currently, state estimation is performed with dead-reckoning, assuming no-slip conditions on the stance feet. The lack of a localization scheme, however, can lead to uncorrected state estimation drift. As discussed, this led to two robot falls in which the segmented planar regions drifted in the positive z direction. State estimation drift also reduces the efficiency of manipulation tasks, requiring multiple realignments of the ATs. As such, a localization scheme appropriate for humanoids (e.g. \cite{fallon2014drift}) is currently being explored.

\textbf{Improving Autonomy with Generic Affordance Templates, Automatic Stance Generation and Environment Segmentation.}
The current implementation of ATs using spatial waypoints for manipulation is best suited for primitive shapes and known objects such as the TCV and hand tools. However, in unstructured environments ATs may need to have additional parameters that are automatically updated as the object is manipulated. For example, online updating of an object's impedance or its articulation parameters %\cite{arduengo2019versatile} 
would allow for a more generic solution. Combining generic ATs with automatic detection of possible grasps \cite{ten2017grasp} might also improve operator throughput. Additionally this feasibility study reinforces the importance of identifying candidate stance locations for manipulation tasks \cite{james2015prophetic, yang2016idrm}. Future autonomous stance generation would preferably reason about collisions, manipulability, maximum exertable forces, and  contact constraints. In general, it is hypothesized that more onboard autonomy will improve operator and robot performance.

\textbf{Enabling Real-time Waypoint Navigation with Whole-body Awareness:}
The current A* footstep planner returns valid plans using particular assumptions. While the planner is able to find plans within 40s even in highly complex environments, some plans need additional operator inspection before execution due to current limitations discussed in Sec.~\ref{sec:execution_approach_and_discussion}. 

Obtaining more real-time plans will enable online replanning under rapid environment and perceptual changes as well as increase locomotion efficiency in time constrained tasks. The incorporation of whole-body awareness in plans will also increase execution safety and reduce operator load. In general however, improving the planner capability increases complexity and the required planning time. Thus, identifying the proper abstraction to obtain safe and real-time plans remains ongoing work. 

\textbf{Addressing Hardware Issues and Increasing Capability.}
As mentioned, 6 of the 10 fall-related failures experienced during the feasibility runs were caused by incorrect readings from Valkyrie's modified (F/T) sensors. Temperature and load direction sensitivity can both cause biased measurements, leading to incorrect detection of ground contacts. These errors, in turn, can degrade robot state estimation and affect swing trajectories, leading to falls. Future robot design improvements addressing this F/T sensor limitation promise to significantly improve overall performance.

Robustness and reliability of the physical robot and its overall task performance will also benefit from a number of other hardware design improvements in the future.  Stronger, more impact resistant wrists will enhance multi-contact performance.  Higher power density actuation would enable more dynamic locomotion trajectories.  And feedback from additional sensors and cameras could be leveraged to improve remote operator situational awareness.  The results of this feasibility study will serve as a valuable resource during the design of future Valkyrie iterations.

\section{CONCLUDING REMARKS}
Looking forward, humanoid robots represent a valuable tool for the bomb disposal community.  This feasibility study established a viable baseline of performance with the successful completion of relevant tasks by NASA's Valkyrie humanoid.  Fall robustness and recovery remain a significant challenge for eventual field deployment, where even infrequent issues can jeopardize overall mission success.  And time critical tasks will certainly require increases in operator throughput and additional onboard autonomy.  However, the described lessons learned and capabilities demonstrated provide a strong foundation for continued progress toward humanoid robot use in bomb disposal applications.

%for instance by improving controllers and trajectory planners to exploit further dynamic wholebody motions. Multi-contact locomotion and manipulation  for navigating and manipulating in tight spaces is a nice-to-have.}

%\section*{ACKNOWLEDGMENT}
%The authors are grateful to the many supporting engineers and personnel at NASA, TRACLabs, and IHMC.

%%%%%%%%%%%%%%%%%%%%%%%%%%%%%%%%%%%%%%%%%%%%%%%%%%%%%%%%%%%%%%%%%%%%%%%%%%%%%%%%
\bibliographystyle{IEEEtran}
\bibliography{references}

\end{document}